# Detection and Forecasting of Parkinson Disease Progression from Speech Signal Features Using Multi-Layer Perceptron and LSTM

Majid Ali[1], Hina Shakir[2], Asia Samreen[3], Sohaib Ahmed[2]

[1] *Department of Electrical Engineering, Bahria University Karachi Pakistan*

[2] *Department of Software Engineering, Bahria University Karachi Pakistan*

[3] *Department of Computer Science, Bahria University Karachi Pakistan*

**Abstract**

Accurate diagnosis of Parkinson disease, especially in its early stages, can be a challenging task. The application of machine learning techniques helps improve the diagnostic accuracy of Parkinson's disease detection but only few studies have presented work towards the prediction of disease progression. In this research work, Long Short-Term Memory(LSTM) was trained using the diagnostic features on Parkinson patients speech signals, to predict the disease progression while a Multilayer Perceptron(MLP) was trained on the same diagnostic features to detect the disease. Diagnostic features selected using two well-known feature selection methods named Relief-F and Sequential Forward Selection and applied on LSTM and MLP have shown to accurately predict the disease progression as stage 2 and 3 and its existence respectively.

*Keywords:* Parkinson Disease detection, Diagnostic features, disease progression

## 1. Introduction

Parkinson disease (PD) is a progressive and degenerative illness that affects the nervous system and impairs movement control [1, 2]. It typically affects around one percentage of the population over the age of 60, with an occurrence rate of approximately 250 individuals per 100,000 people [3, 4]. While signs and symptoms can vary from one patient to another, common speech-related symptoms for Parkinson's disease patients include reduced volume of speech, a monotonous pitch, changes in voice quality, and abnormally fast speech, often referred to as hypokinetic dysarthria. People with Parkinson's disease may not realize that they are speaking softly, so others often ask them to speak louder [4].

Approximately 90% of individuals with PD also experience some form of speech difficulty. This has led to a growing interest in utilizing voice measurements to detect and monitor the symptoms of PD. While physical conditions such as vocal nodules, vocal cord paralysis after a stroke or surgery, or contact ulcers on

Corresponding author : Hina Shakir
Email: hinashakir.bukc@bahria.edu.pk

the vocal cords can contribute to voice disorders, these issues can also arise from vocal misuse, such as speaking too high or low in pitch, too softly or loudly, or with inadequate breath support, often due to postural problems [5]. Typically, predictions are based on clinical practices and neurological examinations, which involve assessing the patient in person using a novel scoring system known as Unified Parkinson's Disease Rating Scale (UPDRS). UPDRS stands as an invaluable tool in the realm of PD assessment, offering a comprehensive framework to evaluate a diverse array of symptoms. The UPDRS assessment consists of four parts: Part I evaluates non-motor symptoms, Part II assesses activities of daily living, Part III evaluates motor symptoms, and Part IV examines treatment complications.

There is an increasing acknowledgment of the importance of vocal signals within the domain of ML-based voice analysis. Failing to detect PD in its early stages can lead to severe and even fatal consequences. Timely intervention is crucial for improving the quality of life for affected individuals [6]. The accurate and timely diagnosis of PD remains a challenge, necessitating innovative approaches to enhance diagnostic accuracy [1, 7]. Machine learning (ML) approaches have demonstrated their reliability as a diagnostic tool, in the context of conventional approaches using chemical, physiological, or electrical inputs.

The objective of this research work is to investigate novel features from the speech samples and integrate those features in machine learning models to obtain accurate Parkinson's disease stage diagnosis and progression prediction.

*1.1. Related Work*

In recent years, extensive research has been conducted to diagnose PD using voice signals and various machine learning techniques. Lahmiri et al. [8] implemented a diverse set of eight feature selection techniques to reduce dataset dimensionality including t-test, entropy, ROC, Bhattacharyya statistics, Wilcoxon statistics, Fuzzy Mutual Information, Genetic Algorithm, and Recursive Feature Elimination with SVM Correlation Bias Reduction. The results exhibited high sensitivity and specificity in achieving accurate and reliable detection of Parkinson's disease. Braga& Ajith et al. [9] proposed early detection of PD, focusing on free-speech recordings captured in uncontrolled background conditions and their detection mechanism integrated signal and speech processing techniques with ML algorithms.

Despotovic et al [10] the study focused on enhancing feature selection efficiency in PD detection by combining Gaussian process with the Automatic Relevance Determination (ARD) feature selection technique. Haq et al.[11] proposed a PD prediction system utilizing a SVM as the predictive model. The authors incorporated an L1-Norm SVM for feature selection with cross validation technique, ensuring accurate

Corresponding author : Hina Shakir
Email: hinashakir.bukc@bahria.edu.pk

classification of PD and healthy control subjects. Tuncer et al. [12] proposed a novel combination of Minimum Average Maximum tree and Singular Value Decomposition as a feature extraction technique for PD diagnosis. Rizvi, Danish et al. [13] researched and investigated a predictive model for Parkinson's disease, employing a deep neural network (DNN) and a long short-term memory (LSTM) network-based approach with voice samples. Hadeel Ahmed et al. [14] implements a recurrent neural network with LSTM, integrating batch normalization and the adaptive moment estimation (ADAM) optimization algorithm to improve PD classification performance. Rohit Lamba et al. [15] developed a hybrid model using machine learning techniques and feature extraction from PD and healthy patients voice data for disease detection.

### 1.2. *Significance of the Proposed Research Work*

This research distinguishes itself from prior studies [8]-[15] by exploration of feature selection techniques combined with machine learning models for stage diagnosis as well as prediction of Parkinson's disease progression. Unlike the previous research work which focused on disease detection only, the presented research work evaluates MLP, SVM and LSTM with advanced feature selection methods like Relief-F and Sequential Forward Selection (SFS) on the Motor-UPDRS to offer stage diagnosis and prediction both. The LSTM-based Recurrent Neural Network model has not been examined using novel features selected for PD forecasting.

The remaining paper comprises of three sections. In Section two, the proposed methodology and architecture of employed machine learning models are presented. In Section three, results are furnished and discussion is done whereas conclusion of the research work is given in Section four.

### 2. Methods/Experimental

The work flow of the proposed research process is given in Fig. 1. The research approach involves several sequential steps to develop a model for diagnosing as well as predicting the progression of PD. The first step is data acquisition, where the necessary dataset for analysis is gathered. Once the data is collected, a pre-processing technique, specifically normalization, is applied to standardize the data and remove any inconsistencies followed by class balancing using Synthetic Minority Oversampling Technique (SMOTE). Next, feature extraction techniques namely Relief-F and Sequential Forward Selection are utilized to identify and select the most relevant features from the dataset. These techniques aid in decreasing the dimensionality of the data, allowing a concentration on the most informative features. After feature extraction, machine learning models had been constructed for this regression task related to the table diagnosis and prediction of PD stages. Both MLP and SVM, which are effective algorithms for analysis are

Corresponding author : Hina Shakir
Email: hinashakir.bukc@bahria.edu.pk

used in this study. The constructed machine learning models are then trained using the pre-processed dataset. Training involves adjusting the model's parameters to optimize its performance and accuracy. Once the models are trained, these can be employed for stage diagnosis and prediction, providing estimates of continuous numerical values related to the research problem.

Corresponding author : Hina Shakir
Email: hinashakir.bukc@bahria.edu.pk

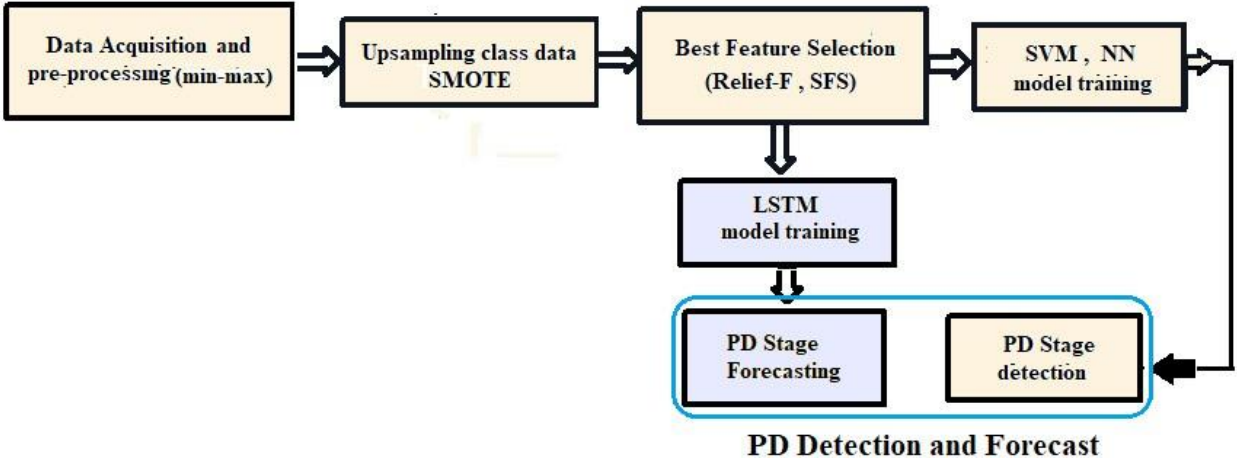

Figure 1: Work Flow of the Proposed Research Work

*2.1. Data Collection and Analysis*

The initial step of this study involves gathering voice data from a public database for analysis. The data collected includes the UPDRS assessment, which is a parameter used for evaluating the motor symptoms of PD. The assessment is conducted by a movement disorder specialist. The Parkinson's tele−monitoring voice dataset from the UCI Machine Learning Repository [16] is used, which includes voice measurements of 42 patients. Each patient has around 200 recordings. These participants were enrolled in a six-month trial for evaluating a tele−monitoring device, which aimed to remotely monitor the progression of symptoms. The recordings were automatically captured within the patients' homes as part of the trial. This dataset has multivariate characteristics and includes 5,875 instances. The dataset provided is predominantly utilized for regression analysis purposes. Table 1 presents comprehensive information regarding the attributes of the data. The 70% of each group data about 4114 instances are used for training purposes, while 30% of each group about 1761 instances are allocated for testing purposes. Among these attributes, motor-UPDRS and Total-UPDRS are the output variables with different ranges corresponding to 4 different PD stages, whereas all other features serve as inputs to the machine learning model. Fig.2 illustrates the distribution of data among these stages, reflecting the allocation of individual data points. These stages are standard in clinical assessments for categorizing individuals with PD [17].

Corresponding author : Hina Shakir
Email: hinashakir.bukc@bahria.edu.pk
5

Table 1: Dataset Attribute Information

| Attribute | Description |
|---|---|
| Subject Number | An integer serving as a unique identifier for each subject |
| Age | The age of the subject |
| Sex | The gender of the subject, |
| Test time | The duration since the subject's recruitment in the trial (in seconds) |
| Motor UPDRS (output) | The clinician's Motor-UPDRS score |
| Total UPDRS (output) | The clinician's Total-UPDRS score |
| Jitter: Discrete Difference Pitch(DDP) | Measures quantifying the variations in frequency |
| Jitter (%) | |
| Jitter: Relative Average Perturbation (RAP) | |
| Jitter: (Absolute) | |
| Jitter: Pitch Period Perturbation Quotient 5 (PPQ5) | |
| Shimmer: Discrete Difference Amplitude(DDA) | Measures quantifying the variations in amplitude |
| Shimmer: Decibel (DB) | |
| Shimmer: Amplitude Perturbation Quotient 5(APQ5) | |
| Shimmer: Amplitude Perturbation Quotient 11(APQ11) | |
| Shimmer: Amplitude Perturbation Quotient 3 (APQ3) | |
| Normalized High-frequency Power Ratio (NHR) | Measures representing the ratio of noise to total components |
| Harmonic-to-Noise Ratio (HNR) | |
| Recurrence Plot Density Entropy (RPDE) | A nonlinear dynamical complexity measure |
| Detrended Fluctuation Analysis (DFA) | A signal fractal scaling exponent |
| Pitch Period Entropy (PPE) | A measure to record the fluctuation in fundamental frequency |

Corresponding author : Hina Shakir
Email: hinashakir.bukc@bahria.edu.pk

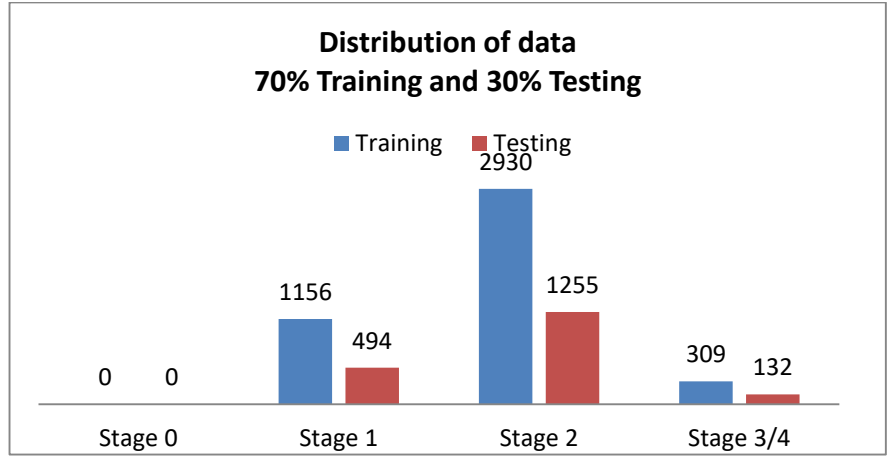

Figure 2: Group-wise Distribution of Experimental Dataset

### 2.2. Data Pre−processing (SMOTE and Data Normalization)

The class imbalance in the UCI in experimental data set was addressed using Synthetic Minority Oversampling Technique (SMOTE) to create new instances of minority class sample. It starts by selecting a sample and its nearest neighbors and then constructs synthetic samples by interpolating between the selected sample and its neighbors. By utilizing SMOTE, the number of stage 3/4 samples has been significantly increased, expanding the dataset from 41 to 341 instances.

After class balancing, the data was normalized using min-max and z-score normalization and tested on the proposed model. The min-max normalization technique exhibited higher accuracy than *z-score* normalization and. Furthermore, pre−processing steps involve removing any missing values or outliers from the data.

### 2.3. Feature Selection Methods

For the diagnosis of Parkinson's disease, two well-known feature selection algorithms, Relief-F algorithm and Sequential Forward Selection had been implemented. These algorithms are well-known to identify and choose the most pertinent features, contributing to the precise and efficient diagnosis and predict the progression of PD. The Relief-F evaluates the relevance of features by considering their ability to discriminate between different classes. It measures the importance of each feature based on the difference between the feature values of the nearest instances of the same and different classes. By assigning weights to the features based on their significance, the algorithm identifies the most discriminative ones, which are crucial for accurate classification.

The Sequential Forward Selection (SFS) is a feature selection technique that aims to build an optimal subset of features by iteratively adding one feature at a time. It starts with an empty set of features and

Corresponding author : Hina Shakir
Email: hinashakir.bukc@bahria.edu.pk

incrementally selects the most informative feature in each iteration that maximizes the improvement in the model's performance[18-20]. This process continues until a specified number of features or a predefined performance criterion is met. It explores the space of feature combinations, potentially discovering synergistic effects between features that contribute to better classification performance [21]. The top ten ranked features using the RELIEF-F and SFS techniques are reported in Table 2. There are seven features which are common in the top ranking features list developed using both types of features.

*2.4. Building Machine Learning Models*

Machine learning algorithms can help uncover hidden patterns and understand the intricate interactions that contribute to the PD. To develop a model for diagnosing and predicting PD, ML algorithms including MLP, SVM, RNN with LSTM had been explored for detection and forecasting of progression of PD.

*2.4.1. MultiLayer Perceptron*

A multilayer perceptron(MLP) is a computational model that mimics the structure and functionality of the human brain. It comprises interconnected neurons arranged in layers, including an input layer, one or more hidden layers, and an output layer. The number of nodes in the hidden layer(s) can be adjusted. The neural network learns from data by adjusting the weights and biases associated with the connections between neurons. This learning process is achieved through an iterative optimization technique called backpropagation, where the model updates these parameters to minimize the difference between predicted and actual target values. Each neuron applies an activation function which is a crucial component in MLP that introduces non-linearity to the model's computations. Regularization techniques like L1 or L2 regularization can be applied to prevent over-fitting. Training a neural network involves providing labeled datasets for supervised learning. The network learns from these examples and becomes capable of processing unknown inputs more accurately [22]. In Fig. 3(a), the proposed framework of a neural network is demonstrated which comprises multiple layers, each with a designated function. These layers collaborate to convert input data into meaningful output predictions.

*2.4.2. Support Vector Machine*

Support Vector Machine (SVM) is a powerful ML model and its purpose is to identify an optimal hyper-plane that distinctly separates data points belonging to different classes or, in the case of regression that best fits the input data. Within SVM, data points are encoded as vectors in a high-dimensional space. The goal is to locate the hyper-plane that most effectively segregates the data points into distinct classes or approximates the regression line. The optimal hyper-plane, known as the decision boundary guarantees the maximum margin, signifying the distance between the hyper-plane and the nearest data points from each class or the

Corresponding author : Hina Shakir
Email: hinashakir.bukc@bahria.edu.pk

deviations from the exact regression line [23].

Corresponding author : Hina Shakir
Email: hinashakir.bukc@bahria.edu.pk

Table 2: List of Selected Features using SFS and RELIEF-F Algorithm

| S no. | SFS Algorithm | RELIEF-F Algorithm |
|---|---|---|
| 1 | Total UPDRS | Test time |
| 2 | Subject Number | Total UPDRS |
| 3 | Age | Subject Number |
| 4 | Test Time | Age |
| 5 | Sex | NHR |
| 6 | Jitter (%) | Sex |
| 7 | Jitter (Abs) | Jitter (Abs) |
| 8 | Jitter: DDP | DFA |
| 9 | Shimmer: APQ5 | Jitter: RAP |
| 10 | Jitter: PPQ5 | Jitter: DDP |

*2.4.3. Long Short-Term Memory (LSTM)*

A Long Short-Term Memory (LSTM) model is a type of Recurrent Neural Network and is capable of learning long-term dependencies and sequential patterns and enhances prediction accuracy. The LSTM in the presented research work was employed for forecasting Stage 2 and Stage 3 respectively based on Stage 1 and Stage 2 data. The model consists of sequence input layer, an LSTM layer with a specified number of hidden units, a fully connected layer, and a regression layer.

The LSTM layer is configured with *tanh* activation for the memory cell known as state activation and sigmoid activations for the input, forget, and output gates known as Gate activation. The LSTM functioning involves initialization with the LSTM initializing with a memory cell and a hidden state. For each time step, the LSTM receives an input and the previous hidden state. The input is used to compute values for the input, forget, and output gates. These values are passed through activation functions (sigmoid) to produce gate values between 0 and 1. The cell state is updated using the input and forget gates to decide what information to store or discard. The hidden state is updated based on the cell state and the output gate. The updated hidden state is used for making predictions. The output is produced based on the hidden state and can be used for tasks such as classification or regression. The proposed framework of LSTM used in the research work is illustrated in Fig. 3(b).

Corresponding author : Hina Shakir
Email: hinashakir.bukc@bahria.edu.pk

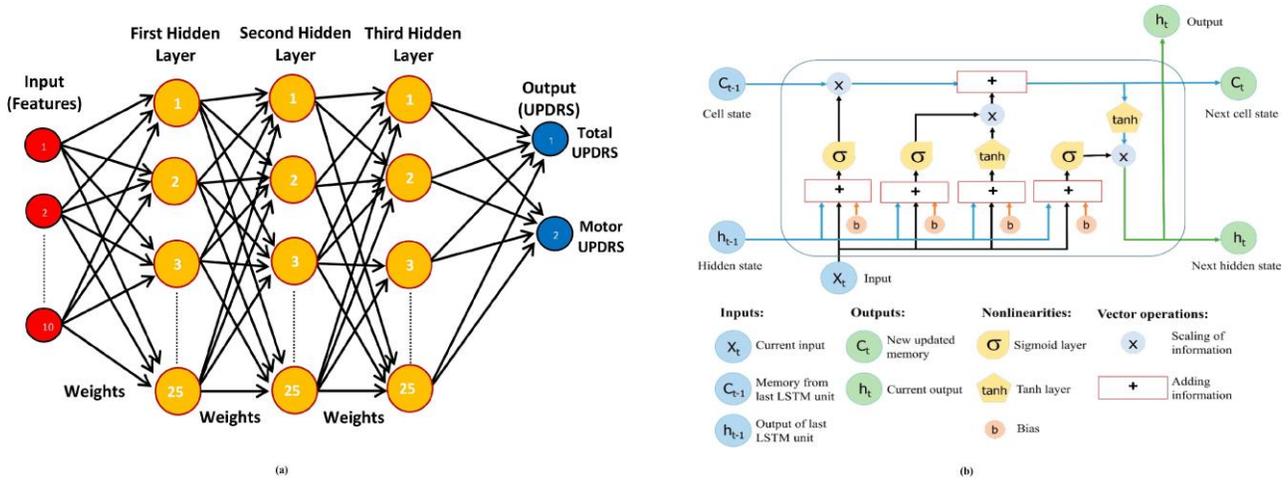

Figure 3: Proposed Framework of Machine Learning Models including (a) MLP (b)LSTM

The model consists of sequence input layer, an LSTM layer with a specified number of hidden units, a fully connected layer, and a regression layer. The LSTM layer is configured with tanh activation for the memory cell known as state activation and sigmoid activations for the input, forget, and output gates known as Gate activation. The LSTM is initialized with a memory cell and a hidden state. For each time step, the LSTM receives an input and the previous hidden state. The input is used to compute values for the input, forget, and output gates. These values are passed through activation functions (sigmoid) to produce gate values between 0 and 1. The cell state is updated using the input and forget gates to decide what information to store or discard. The hidden state is updated based on the cell state and the output gate. The updated hidden state is used for making predictions.

*2.4.4. Parameter tuning of ML Models*

During the research work, several adjustments were made to the MLP, SVM and LSTM models to optimize their performance and the tuned parameters for each of the models are reported in Table 3. The neural network comprises three fully connected layers, each containing 25 nodes. The Rectified Linear Unit (ReLU) activation function had been employed as the activation function. To prevent the training algorithm from running indefinitely, an iteration limit of 1000 is set. In this study, no regularization is applied to the network as the regularization strength (Lambda) is set to 0 and a cross-validation of 10-fold cross-validation had been employed. The Limited-memory Broyden-Fletcher-Goldfarb-Shanno (LBFGS) algorithm is employed as an optimization technique designed for unconstrained optimization problems. It falls within the category of quasi-Newton methods and is notably effective for addressing problems characterized by a

Corresponding author : Hina Shakir
Email: hinashakir.bukc@bahria.edu.pk

substantial number of variables.

Corresponding author : Hina Shakir
Email: hinashakir.bukc@bahria.edu.pk

Table 3: Parameter tuning of ML models

| Neural Network Model | |
|---|---|
| **Layer Configuration** | 3 Fully Connected Layers |
| **Number of Nodes** | First Layer: 25 <br> Second Layer: 25 <br> Third Layer: 25 |
| **Activation Function** | ReLU |
| **Iteration Limit** | 1000 |
| **Regularization Strength** | 0 |
| **Data Standardization** | Applied |
| **Learning Rate** | 0.1 |
| **Optimizer** | LBFGS |
| **Cross-validation** | 10-fold |
| **SVM Model Details** | |
| **Model Type** | Gaussian SVM |
| **Kernel Function** | Gaussian |
| **Kernel Scale** | 2 |
| **Box Constraint (C)** | Automatic |
| **Epsilon Value** | Automatic |
| **Data Standardization** | Applied |
| **Optimizer** | SMO |
| **Cross-validation** | 10-fold |
| **LSTM Model** | |
| **Model Type** | Sequential LSTM |
| **Layer Configuration** | Sequence Input Layer (1x1) <br> LSTM Layer (1x1) <br> Fully Connected Layer (1x1) <br> Regression Output Layer (1x1) |
| **State Activation Function** | Tanh |
| **Gate Activation Function** | Sigmoid |
| **Optimizer** | Adam |
| **Maximum Epochs** | 1000 |
| **Validation Frequency** | 50 |
| **Number of Hidden Units** | 150 |
| **Learning Rate** | 0.001 |

Corresponding author : Hina Shakir
Email: hinashakir.bukc@bahria.edu.pk

For the SVM, a Gaussian Radial Basis Function (RBF) kernel function is used which is capable of handling non−linear relationships by mapping input features into higher-dimensional space. The Kernel Scale is set to 2 after testing with 0, 1, 2, 3 and 4. The epsilon value is set to 'Automatic' and is primarily relevant in SVM regression tasks and defines the width of the epsilon−insensitive zone. Data standardization (Normalization) and Sequential Minimal Optimization (SMO) is applied which efficiently solve the quadratic programming (QP) problem that arises during the training of SVM. A cross−validation of 10−fold was used.

The Sequential LSTM-based Regression Neural Network employs Sequence Input Layer, LSTM Layer, Fully Connected Layer and Regression Output Layer with each having configuration of 1x1. The state activation function is set to the hyperbolic tangent (tanh) function with values between -1 and 1. The gate activation functions (Input Gate, Forget Gate, and Output Gate) are set to the sigmoid function with the Adam optimizer chosen as optimization algorithm for training. The maximum number of training epochs is set to 1000 and validation is performed every 50 iterations or mini-batches. The number of hidden units is 150 with learning rate fixed at 0.001 throughout training.

*2.4.5. Performance Evaluation Metric*

To assess the prediction performance of each method; MSE and R-Squared were used. MSE provide insights into the magnitude of errors and R-squared, on the other hand, indicates the goodness of fit of the model by revealing how well the independent variables explain the variability in the dependent variable [24].

The formulae to compute R-squared and MSE are given below :

$$\text{R} - \text{squared} \ = 1 - \left(\frac{\text{SmS}_{res}}{\text{SmS}_{totl}}\right) \quad (1)$$

Where *SmS$_{res}$* is the sum of the squared differences between the predicted and actual values of y and *SmS$_{totl}$* represents the sum of the squared differences between the actual values of y and their mean. indicating better model performance.

$$\text{MSE} \ = \frac{1}{n} * \sum(y - y_{predicted})^2 \quad (2)$$

Corresponding author : Hina Shakir
Email: hinashakir.bukc@bahria.edu.pk

Table 4: ML Models Performance for PD Stage Detection

|  | Neural Network | | Gaussian SVM | |
|---|---|---|---|---|
|  | MSE | R-Squared | MSE | R-Squared |
| **Without Feature Selection** | | | | |
| **Training** | 0.0590 | 1.00 | 2.361 | 0.97 |
| **Test** | 1.1847 | 0.98 | 5.286 | 0.91 |
| **Relief-F** | | | | |
| **Training** | 0.0311 | 1.00 | 0.8193 | 0.99 |
| **Test** | 0.9745 | 0.98 | 3.3467 | 0.94 |
| **Sequential Forward Selection (SFS)** | | | | |
| **Training** | 0.0358 | 1.00 | 0.8428 | 0.99 |
| **Test** | 0.6275 | 0.99 | 2.6863 | 0.95 |

Where n is the number of observations/instances, y is actual value of the dependent variable and $y_{predicted}$ represents the predicted value of dependent variable.

## 3. Results and Discussion

The experimental findings and discussion of the proposed machine learning approach are presented into two parts : first for PD stage detection; and then for PD stage forecast.

*3.1. PD Detection*

A detailed comparison of performance of the machine learning models (MLP and SVM) for PD detection using features from Relief and SFS algorithms is reported in Table 4.

The Relief-F algorithm and MLP model produced the MSE of 0.9745 on test data with the R-Squared value of 0.98. For Relief-F SVM model, the MSE was 3.3467 and the R-Squared value was 0.94. The performance of models for training data was significantly better for al presented scenarios.

Corresponding author : Hina Shakir
Email: hinashakir.bukc@bahria.edu.pk

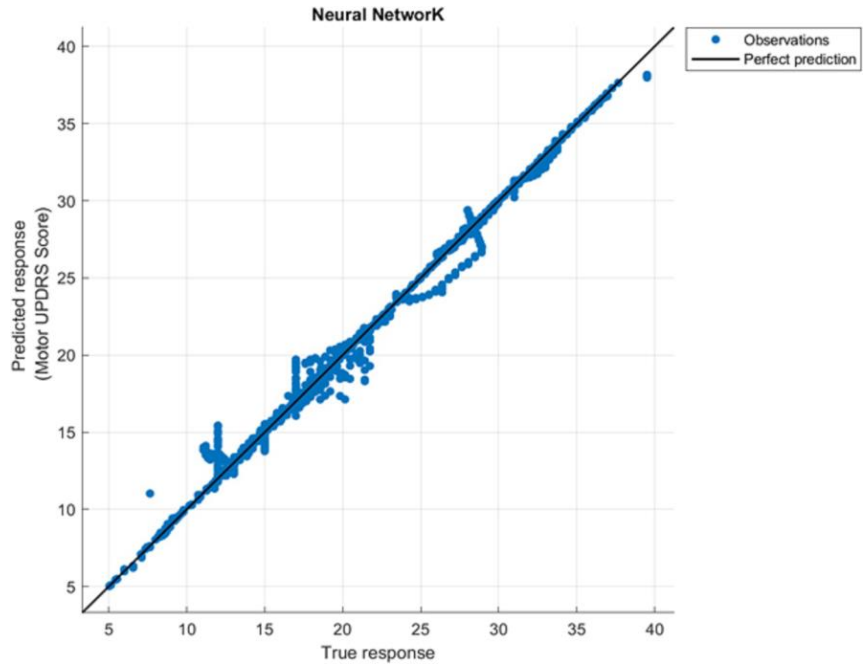

Figure 4: True Responses vs. Predicted Response of MLP−SFS Model

The SFS algorithm with MLP Model resulted in an MSE value of 0.6275 and the R-Squared value of 0.99 for the test data. For the SFS algorithm and Gaussian SVM, MSE was 2.6863, and the R-Squared value is 0.95. For a comparison, both MLP and SVM were evaluated for PD detection without feature selection. The MSE value and R-Squared value were higher in both scenarios than the values obtained when the input was selected features.

Notably, the neural network when used in combination with the SFS feature selection, emerged as the top-performing model. Fig. 4 illustrates the comparison between true responses and predicted responses of the final model, which was built with SFS feature selection method in conjunction with a MLP model. A close alignment between the points on the plot suggests that the model is accurate in its predictions, while deviations may indicate areas where the model can be improved. This outcome underscores the effectiveness of the Neural Network in capturing the underlying patterns in the data, particularly when paired with feature selection methods like SFS.

Selecting MLP−SFS model as the better model for PD stage classification, confusion matrix for PD stage classification is demonstrated in Fig. 5. All the instances of stage 2 are classified correctly. However, 20 instances of stage 1 are classified as stage 2 and 4 instances of stage 3 are classified as stage 2. These results emphasize the importance of feature selection techniques in enhancing accuracy. The MLP−SFS combination shows promise for precise disease management, and these findings provide a strong basis for further research and

Corresponding author : Hina Shakir
Email: hinashakir.bukc@bahria.edu.pk

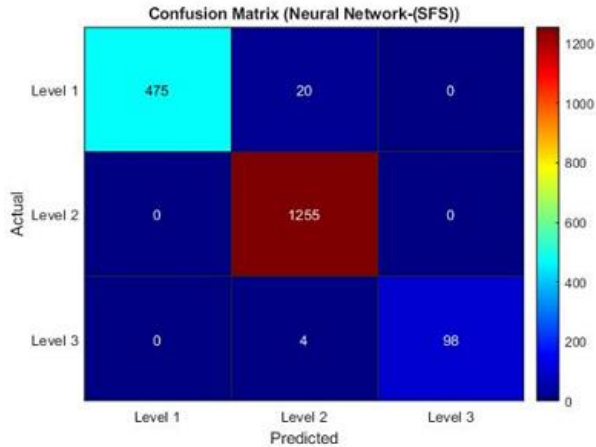

Figure 5: Confusion Matrix of PD stage classification using NN−SFS Model

clinical tools to enhance Parkinson's disease assessment and management.

## 3.2. PD Stage Forecasting

Table 5 provides insights into the results of LSTM Forecast for stage 2 and stage 3 respectively. The training was conducted for 1000 epochs, with updates after each iteration. The results are presented at selected iterations (e.g., 50, 100, 150, 200, etc.). The Root Mean Squared Error (RMSE) serves as a metric for assessing the accuracy of the model's predictions and the validation loss values signify the disparity between predicted and actual values, with lower values indicating a reduction in the gap between predicted and actual values throughout the training process.

The stage 2 prediction over multiple epochs exhibits decreasing trends in error and loss metric suggest effective learning over time. The average RMSE is reported as 0.88, and MSE is 0.77, which indicates good model's performance. The Stage 3 forecast model incorporating information from Stages 1 and 2 achieves an MSE of 3.16 and an RMSE of 1.78 respectively, offering a detailed assessment of accuracy parameter.

The LSTM model demonstrates learning and improved forecasting capabilities across both Stage 2 and Stage 3. The convergence of both training and validation metrics indicates that the model is stabilizing and making accurate predictions. The test data analysis indicates that 1560 samples were accurately forecasted within the specified Stage 2 range. Additionally, 2625 samples were correctly forecasted, albeit with a slight deviation from the Stage 2 range, falling into the Stage 1 range. Similar occurrences are observed for Stage 3 forecasting, where, due to the lower number of test samples, Stage 3 faces comparable challenges.

Corresponding author : Hina Shakir
Email: hinashakir.bukc@bahria.edu.pk

Table 5: LSTM Forecast Performance of PD Stages

| PD Stage 2 Forecast | | | PD Stage 3 Forecast | | |
|---|---|---|---|---|---|
| **Epoch** | **Validation RMSE** | **Validation Loss** | **Epoch** | **Validation RMSE** | **Validation Loss** |
| 1 | 12.00 | 71.9485 | 1 | 21.77 | 236.8813 |
| 100 | 2.67 | 3.5727 | 100 | 7.51 | 28.1740 |
| 200 | 1.75 | 1.5258 | 200 | 3.42 | 5.8389 |
| 400 | 0.96 | 0.4565 | 400 | 2.00 | 2.0040 |
| 600 | 0.97 | 0.4681 | 600 | 1.97 | 1.9437 |
| 800 | 0.89 | 0.3991 | 800 | 1.62 | 1.3103 |
| 1000 | 0.88 | 0.3950 | 1000 | 1.78 | 1.5793 |

Finally, the result of our proposed method is compared with significant research methods presented using the same UCI Parkinson's dataset. Table 6 draws a comparison of the results of this research with three other prominent works in terms of accuracy and output The proposed work is novel as the other three methods have only proposed PD detection models using ML methods , while the proposed method also performs PD stage forecasting. Fatlawi et al [25], Rasheed et al. [26] and Alshammri et al. [27] obtained accuracy of 94%, 97.50% and 98.31% in PD detection using Data Belief Networks(DBN), Back propagation Variable Adaptive Momentum(BPVAM) and Multilayer Perceptron (MLP) whereas the proposed NN-SFS model achieved an accuracy of 98.63%, exceeding the performance of the previous research methods. In addition, the presented LSTM with SFS secured an 88.7% accuracy in PD forecasting which can be improved by providing more samples of Stage 3. The comparison shows the superior performance of the presented work.

Corresponding author : Hina Shakir
Email: hinashakir.bukc@bahria.edu.pk

Table 6: Comparison of Proposed Model with State of the Art Techniques

| Article(year) | Technique | Outcome | Performance |
|---|---|---|---|
| Fatlawi et al.(2016) | DBN | PD detection | Acc. =94% |
| Rasheed et al.(2020) | BPVAM-PCA | PD detection | Acc. =97.50% |
| Alshammri et al.(2023) | MLP | PD detection | Acc. =98.31% |
| Proposed method | SFS -MLP | PD detection | Acc. 98.63% |
|  |  | PD forecasting | Acc. =88% |

## 4. Conclusions

This research work provides valuable insights into the application of machine learning models along with feature selection methods for assessing the diagnosis and progression of Parkinson's disease through the prediction of Motor UPDRS scores. Noteworthy is the proven robustness and promise of the neural network model, especially when combined with the SFS feature-selecting technique. Furthermore, the study highlights the success of an LSTM in accurately forecasting Stage 2 and Stage 3 data. The commendable RMSE values underscore the effectiveness of the RNN-LSTM model in making precise and reliable predictions. These findings carry significant implications for advancing our understanding of Parkinson's disease and its progression, facilitating early detection, and tailoring treatment strategies. Exploring alternative feature extraction techniques, such as Perception Linear Predictive Coefficients (PLP) or wavelet transforms, is proposed to assess their potential for enhancing model performance. Another avenue is the integration of speech analysis with other biomarkers, including genetic data and neuroimaging, to create more accurate and reliable machine learning models for Parkinson's disease diagnosis and progression monitoring.

**List of Abbreviations**

```
PD      -   Parkinson's Disease
NN      -   Neural Network
SVM     -   Support Vector Machine
UPDRS   -   Unified Parkinson's Disease Rating Scale
ML      -   Machine Learning
SFS     -   Sequential Forward Selection
ROC     -   Receiver Operating Characteristic
PCA     -   Principal Component Analysis
DL      -   Deep Learning
UCI     -   University of California, Irvine
Relu    -   Rectified Linear Unit
Tanh    -   Hyperbolic Tangent
```

Corresponding author : Hina Shakir
Email: hinashakir.bukc@bahria.edu.pk

| | | |
|---|---|---|
| LSTM | - | A Long Short-Term Memory |
| RNN | - | Recurrent Neural Network |
| MSE | - | Mean Squared Error |
| RMSE | - | Root Mean Squared Error |
| SMOTE | - | Synthetic Minority Over-sampling Technique |
| RBF | - | Radial Basis Function |
| DBN | - | Data Belief Network |
| MLP | - | Multilayer Perceptron |
| PCA | - | Principal Component Analysis |
| PBVAM | - | Backpropagation Variable Adaptive Momentum |

**Declarations**

- **Availability of data and material**

  The experimental dataset is publicly available.

- **Competing interests**

  The authors have no competing interests to declare that are relevant to the content of this article.

- **Funding**

  Not Applicable

- **Authors' contributions**

  All authors contributed to the study's conception and design. MA performed the data collection, simulation, analysis, and manuscript writing. HS designed the experiments and edited the manuscript. AS helped with the analysis and conducted constructive discussion. All authors read and approved the final manuscript.

- **Acknowledgements**

  Not Applicable

Corresponding author : Hina Shakir
Email: hinashakir.bukc@bahria.edu.pk

Corresponding author : Hina Shakir
Email: hinashakir.bukc@bahria.edu.pk

**This paper is accepted by AGH Computer Science Journal.**

Corresponding author : Hina Shakir
Email: hinashakir.bukc@bahria.edu.pk